\documentclass[runningheads]{llncs}

\usepackage{eccv}

\usepackage{eccvabbrv}

\usepackage{hyperref}

\usepackage{orcidlink}

\usepackage{enumitem}
\usepackage{multirow}
\usepackage{booktabs,multirow,adjustbox,diagbox,threeparttable,tabularray,setspace}
\usepackage{float}
\newcommand{\methodname}{MMControl}
\newcommand{\method}{\texttt{\methodname}\xspace}

\newcommand{\equalcontrib}{\textsuperscript{*}}
\newcommand{\corrauthor}{\textsuperscript{\ensuremath{\dagger}}}

\begin{document}

\title{MMControl: Unified Multi-Modal Control for Joint Audio-Video Generation} 

\titlerunning{Unified Multi-Modal Control for Joint Audio-Video Generation}




\author{
Liyang Li\equalcontrib
\and
Wen Wang\equalcontrib
\and
Canyu Zhao
\and
Tianjian Feng
\and
\\
Zhiyue Zhao
\and
Hao Chen
\and
Chunhua Shen\corrauthor
}

\authorrunning{L.~Li et al.}

\institute{Zhejiang University, Hangzhou, China}

\maketitle

\begingroup
\renewcommand{\thefootnote}{\ifcase\value{footnote}\or *\or \ensuremath{\dagger}\fi}
\footnotetext[1]{Equal contribution.}
\footnotetext[2]{Corresponding author.}
\endgroup


\begin{abstract}

Recent advances in Diffusion Transformers (DiTs) have enabled high-quality joint audio-video generation, producing videos with synchronized audio within a single model. 
However, existing controllable generation frameworks are typically restricted to video-only control. 
This restricts comprehensive controllability and often leads to suboptimal cross-modal alignment. 
To bridge this gap, we present \textbf{MMControl}, which enables users to perform \textbf{M}ulti-\textbf{M}odal \textbf{Control} in joint audio-video generation. MMControl introduces a dual-stream conditional injection mechanism. It incorporates both visual and acoustic control signals---including reference images, reference audio, depth maps, and pose sequences---into a joint generation process. These conditions are injected through bypass branches into a joint audio-video Diffusion Transformer, enabling the model to simultaneously generate identity-consistent video and timbre-consistent audio under structural constraints. Furthermore, we introduce modality-specific guidance scaling, which allows users to independently and dynamically adjust the influence strength of each visual and acoustic condition at inference time. Extensive experiments demonstrate that MMControl achieves fine-grained, composable control over character identity, voice timbre, body pose, and depth-guided scene structure in joint audio-video generation.


\end{abstract}

\begin{figure*}[h]
    \centering
    \includegraphics[width=\textwidth]{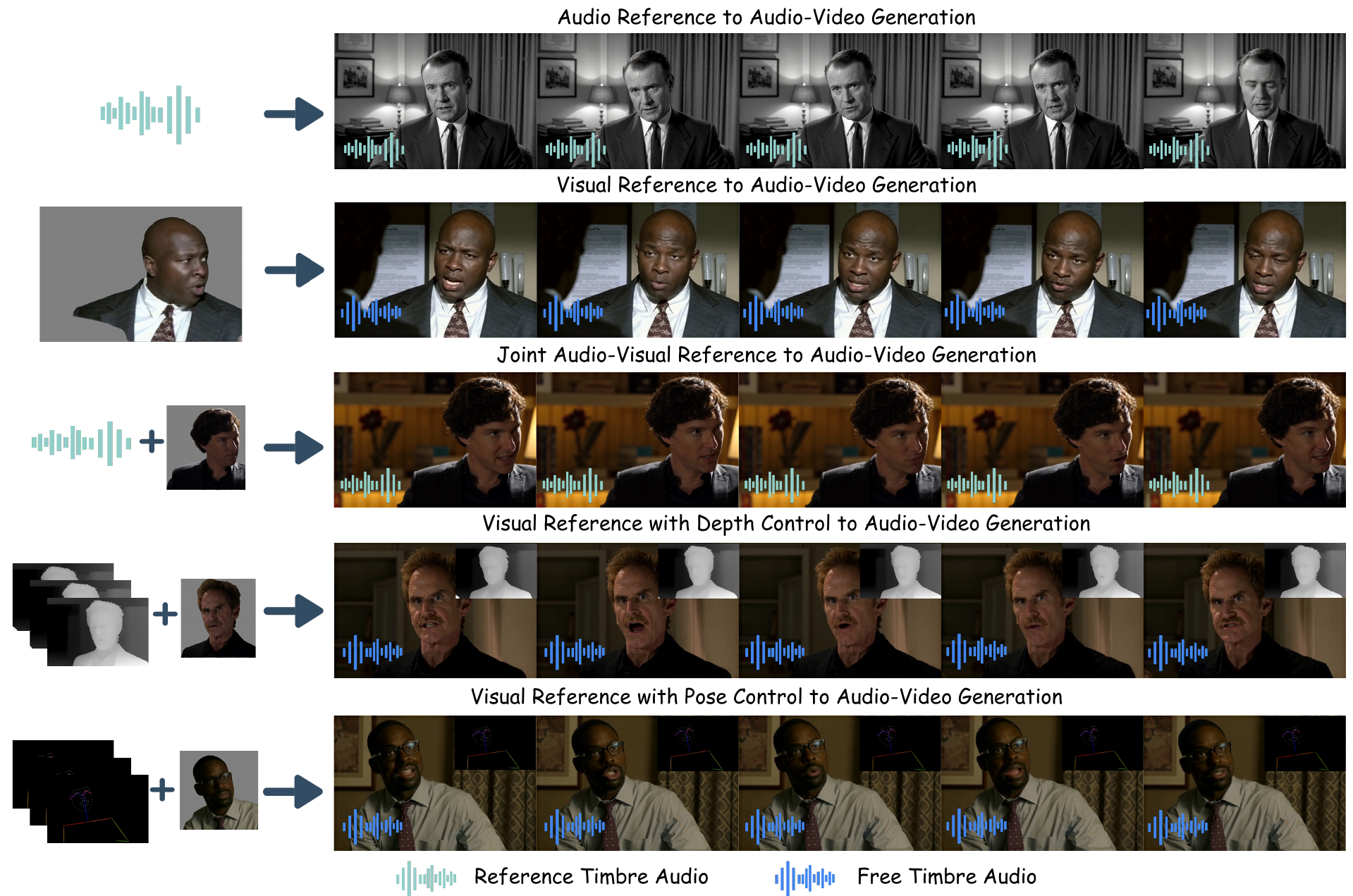}
    \caption{\textbf{Unified Multi-Modal Control results produced by \methodname{}.} 
By providing different control signals, our model enables fine-grained control over joint audio-video synthesis. 
The teaser illustrates our model's ability to generate content that is: 
\textbf{Consistent} in voice and identity (via audio/visual references) and 
\textbf{Controllable} in structure and motion (via depth/pose guidance).
}
    \label{fig:teaser}
\end{figure*}

\section{Introduction}

The pursuit of photorealistic digital content creation has driven generative models to evolve from static image synthesis toward dynamic, high-fidelity video generation. With the advent of Diffusion Transformers (DiTs)~\cite{peebles2023scalable}, visual synthesis has achieved remarkable progress in both scalability and temporal consistency~\cite{kong2024hunyuanvideo, wan2025wan}. More recently, the field has advanced toward a more ambitious frontier: native joint audio-video generation~\cite{openai_sora2, hacohen2026ltx, team2026mova, google_veo3}. Unlike conventional cascaded pipelines, these models learn to synthesize both modalities within a unified latent space, yielding intrinsic synchronization and a more holistic understanding of temporal dynamics.

Despite the impressive generation quality of recent joint DiT models~\cite{team2026mova, hacohen2026ltx}, they still lack fine-grained controllability over generated content. Existing methods are typically either visual-centric~\cite{zhang2023adding, xie2023omnicontrol} or audio-centric~\cite{tian2024emo, wei2024aniportrait, hu2025hunyuancustom}. Visual-centric approaches primarily optimize visual control while under-modeling acoustic conditions, whereas audio-centric approaches emphasize speech or audio guidance but often struggle to produce naturally aligned audio-video sequences. These methods rely on decoupled controls across video and audio modalities, which makes coherent cross-modal alignment difficult. At the core lies the absence of a unified control mechanism that can simultaneously handle visual and acoustic conditioning signals. This exposes a critical gap: achieving precise multi-modal control within joint generation frameworks while preserving inherent cross-modal synchronization and collaborative priors remains highly challenging.

In this work, we focus on the task of Multi-Modal Controllable Joint Generation. The goal is to generate synchronized video and audio that follow textual descriptions while being conditioned on multimodal controls for visual structure (depth/pose), character identity, and vocal timbre. To this end, we propose \method, a new pipeline that introduces fine-grained controllability into joint audio-video generation. At the core of our method is a Multi-Modal Control Unit (MMCU), a unified interface that tokenizes heterogeneous inputs including reference images, reference audio clips, and depth/pose sequences, and aligns them into synchronized representations. This enables \method to handle these control scenarios under a single, consistent paradigm.

To integrate multimodal signals without compromising the pretrained generation capability of the backbone, we design a \textbf{Dual-Stream Bypass} architecture. This design introduces parallel, modality-specific trainable branches interleaved with frozen joint DiT layers. These branches extract contextual cues from the MMCU and inject them into the main diffusion process through gated residual connections. In addition, we introduce \textbf{Modality-Specific Guidance Scaling} at inference time, allowing users to independently adjust the influence of visual and acoustic controls. This flexibility enables \method to satisfy diverse creative requirements, ranging from strict structural adherence to expressive, identity-consistent synthesis.
Extensive experiments demonstrate that \method establishes a new state-of-the-art in controllable joint generation, delivering exceptional visual fidelity, robust audio-video consistency, and precise structural alignment across reference-guided audio-video generation and depth/pose-controlled structural generation. Our main contributions are summarized as follows:

\begin{itemize}
    \item We define and study the task of multi-modal controllable joint generation, providing a formal framework for jointly controlling identity, timbre, and depth/pose structure within a unified generative space.
    \item We propose \method, which introduces the \textbf{Multi-Modal Control Unit (MMCU)} to effectively unify and synchronize heterogeneous control signals as a versatile interface for joint diffusion.
    \item We design the \textbf{Dual-Stream Bypass architecture} for non-intrusive control-feature injection into frozen joint DiT layers, and further introduce Modality-Specific Guidance Scaling for flexible inference-time adjustment.
\end{itemize}

\section{Related Work}

\noindent\textbf{Audio-Video Generation.}
Building on the success of image diffusion models \cite{ho2020denoising, rombach2022high}, video synthesis has evolved from foundational UNet-based architectures~\cite{ho2022video,blattmann2023align,blattmann2023stable} to highly scalable DiT architectures \cite{peebles2023scalable}. Recent visual foundation frameworks~\cite{yang2024cogvideox, kong2024hunyuanvideo, wu2025hunyuanvideo, wan2025wan} pushed the boundaries of video synthesis. 
Beyond text-to-video synthesis, a significant body of work explores steering visual content via acoustic signals. This is most prominent in portrait animation, where pioneering methods like SadTalker \cite{zhang2023sadtalker} utilized 3D morphable models for motion mapping, while more recent diffusion-based methods \cite{tian2024emo, wei2024aniportrait,xu2024hallo, cui2024hallo2, cui2025hallo3, shen2023difftalk, sun2025vividtalk} have achieved superior expressiveness and lip-sync accuracy. 
More recently, audio-to-video frameworks such as MoCha \cite{wei2025mocha} and Wan2.2-S2V \cite{wan2025wan} have further advanced the synthesis of high-fidelity video from acoustic signals.
A transformative shift has recently occurred toward native joint audio-video generation, where both modalities are synthesized simultaneously within a single, unified framework. Pioneering industrial models such as Sora 2 \cite{openai_sora2} and Veo 3 \cite{google_veo3} have demonstrated that modeling video and audio in a joint latent space leads to superior cross-modal synchronization and temporal realism. This paradigm is further advanced by frameworks like JavisDiT \cite{liu2025javisdit}, LTX-2 \cite{hacohen2026ltx}, and Universe \cite{wang2025universe}.

\noindent\textbf{Controllable Video Generation.} 
The quest for fine-grained controllability, which began with image generation, has remained a persistent mission that continues to drive innovation in the field of video generation.
%
Pioneering works such as ControlNet \cite{zhang2023adding} and T2I-Adapter \cite{mou2024t2i} established the foundational paradigm for spatially-conditioned generation by integrating auxiliary bypass modules to process diverse geometric priors. These frameworks facilitate the precise steering of pretrained diffusion models through the incorporation of structural signals such as Canny edges, depth maps, and human poses. Building upon these advancements, GLIGEN \cite{li2023gligen} and Uni-ControlNet \cite{zhao2023uni} further expanded the scope of controllability to encompass grounded generation and multi-modal conditional fusion.
With the rapid advancement of video generation models, extending spatial control to the temporal domain has emerged as a critical research direction. Recent literature has predominantly focused on adapting image-based conditioning techniques to achieve motion-consistent video synthesis. A major line of work injects diverse structural signals and explicit motion priors (\eg, motion vectors) to guide the global temporal generation process \cite{zhang2024controlvideo, chen2023control, wang2023videocomposer}. Building upon these foundations, subsequent approaches have further advanced toward fine-grained and interactive motion steering, enabling users to precisely dictate dynamic elements such as character poses or localized trajectories \cite{ma2024follow, deng2024dragvideo}.
 Building upon these advancements, recent efforts~\cite{xie2023omnicontrol,jiang2025vace} focus on developing unified frameworks for visual control. 
%
Despite these successes, existing methods primarily focus on the visual modality and do not account for the joint control of audio and video.

\noindent\textbf{Customized Generation.}
Customized generation focuses on the preservation of user-specified subject identities.
\textit{Instance-specific customization} 
achieves high-fidelity identity preservation through dedicated per-subject fine-tuning. These optimization-based approaches effectively propagate subject features across temporal sequences and manage multiple identities via spatial masks and attention-based localization \cite{chen2025multi, stillmoving, wu2025customcrafter, chen2024disenstudio, wang2026customvideo}. 
\textit{End-to-end customization} provides a highly scalable, training-free alternative by integrating identity features through specialized encoder-based conditioning. By leveraging robust feature alignment and localized priors (\eg, facial adapters), these methods seamlessly inject single or multiple identity representations in a single forward pass \cite{ConsisID, idanimator, liu2025phantom, fei2025skyreels, huang2025conceptmaster}, thereby bypassing the need for tedious per-subject tuning.
Despite rapid advancements in visual identity preservation, existing generative frameworks predominantly neglect the acoustic dimension. Specifically, the ability to control a character's unique voice timbre remains an under-explored frontier in joint audio-video generation. In this work, we explicitly address this critical gap.

\section{Method}

\subsection{Task Definition}

We define the task of \textbf{Multi-Modal Controllable Joint Generation}, aiming to synthesize temporally synchronized video $\nu \in \mathbb{R}^{t \times h \times w \times 3}$ and audio $\alpha \in \mathbb{R}^{t \times d}$ guided by a flexible combination of conditional signals. Generation is primarily driven by a textual prompt $T$ following a structured format \texttt{[VISUAL]: $c_v$ [SPEECH]: $c_s$}, where $c_v$ provides the scene's semantic description and $c_s$ specifies the spoken content to facilitate cross-modal semantic alignment. For fine-grained control, the model incorporates optional multi-modal signals: a reference image $I^{\text{ref}}$ defining the character's visual identity, a reference audio clip $A^{\text{ref}}$ providing the target voice timbre, and a sequence of structural constraints $S = \{s_1, s_2, \dots, s_t\}$, instantiated as depth maps or pose sequences in this work. These composable conditions enable versatile scenarios ranging from basic text-to-AV synthesis to reference-guided and structure-guided joint generation. Formally, the process is formulated as:
\begin{equation}
    \nu, \alpha = \mathcal{F}(T, I^{\text{ref}}, A^{\text{ref}}, S),
\end{equation}
where $\mathcal{F}$ is a joint Diffusion Transformer mapping textual and multimodal inputs into temporally aligned streams, effectively ensuring identity consistency, timbre fidelity, and structural adherence to the provided depth or pose sequence.

\subsection{Multi-Modal Control Unit (MMCU)}

To bridge Section 3.1's raw control signals with the joint diffusion process, we introduce the \textbf{Multi-Modal Control Unit (MMCU)}. MMCU transforms diverse inputs into a synchronized representation $\mathcal{U} = \{T, V, A, M_v, M_a\}$, serving as our model's unified interface.

Textual guidance $T$ derives from the described structured prompt. It is encoded into a joint embedding space via a pre-trained text encoder, providing global semantic context that guides visual and acoustic generation through cross-attention.

Visual control signal $V$ and mask $M_v$ are constructed by prepending the reference image latent to the structural guidance sequence. Letting $z_{img} = \text{VAE}(I^{\text{ref}})$ and $z_{s,i} = \text{VAE}(s_i)$ for $i \in \{1, \dots, t\}$, the visual MMCU is formulated as:
\begin{equation}
\begin{aligned}
    V &= \{z_{img}\} + \{z_{s,1}, z_{s,2}, \dots, z_{s,t}\}, \\
    M_v &= \{0\} + \{1\} \times t,
\end{aligned}
\end{equation}
where $+$ denotes temporal concatenation. Binary mask $M_v$ ensures the model preserves reference frame identity ($M_v=0$) while following structural constraints $S$ in generation frames ($M_v=1$). For tasks without structural control, $\{z_{s,i}\}$ is replaced by empty latents $\{\mathbf{0}\} \times t$.

Similarly, acoustic control $A$ and mask $M_a$ align reference timbre with the target generation period. Let $\mathbf{z}_{aud} = \{z_{aud, 1}, z_{aud, 2}, \dots, \\ z_{aud, k}\}$ be the reference audio latent sequence from the Audio VAE, where $k$ is the number of latent tokens for the reference duration. A sequence of silence latents $\mathbf{z}_{sil}$ of length $t$ serves as a placeholder for generation. The acoustic MMCU is formulated as:
\begin{equation}
\begin{aligned}
    A &= \{z_{aud, 1}, \dots, z_{aud, k}\} + \{z_{sil, 1}, \dots, z_{sil, t}\}, \\
    M_a &= \{\mathbf{0}\}_k + \{\mathbf{1}\}_t,
\end{aligned}
\end{equation}
where $\{\mathbf{0}\}_k$ and $\{\mathbf{1}\}_t$ are constant sequences of zeros and ones with lengths $k$ and $t$. This ensures the model observes a reference audio segment to fix the speaker identity before the generative phase. While $\mathbf{z}_{sil}$ currently acts as this placeholder, this formulation allows for future extensions using specific acoustic controls like pitch or energy contours.

\subsection{Architecture}

\begin{figure*}[t]
\centering
\includegraphics[width=1\textwidth]{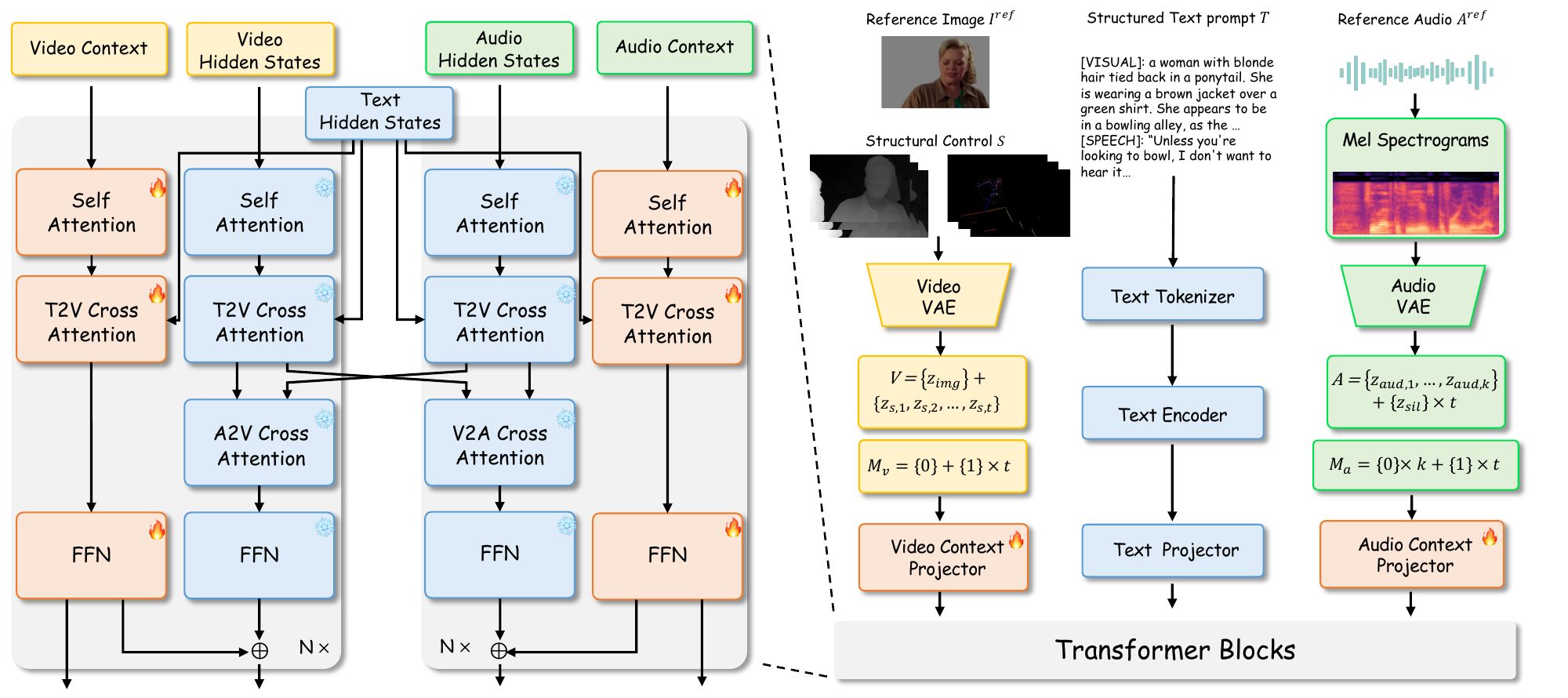}
\caption{
\textbf{Proposed Dual-Stream Bypass architecture for Joint DiT.}
The Dual-Stream Bypass mechanism introduces parallel, trainable visual and acoustic bypass branches interleaved with the frozen Joint DiT backbone at even-numbered layers. 
}
\label{fig:bypass_arch}
\end{figure*}

\noindent \textbf{Dual-Stream Bypass Architecture.} To integrate MMCU signals while preserving pre-trained generative priors, we introduce the Dual-Stream Bypass mechanism. This architecture uses parallel trainable branches to encode multi-modal hints for targeted injection into the frozen Joint DiT backbone for precise conditional control.

\noindent \textbf{Dual-Stream Context Projection.} To effectively integrate MMCU signals, we propose a modality-specific projection mechanism. For each modality $m \in \{v, a\}$, the control sequence $\mathcal{U}_m$ and binary mask $M_m$ are concatenated along the channel dimension:
\begin{equation}
    \mathbf{z}_{in, m} = [\mathcal{U}_m; M_m],
\end{equation}
where $[\cdot ; \cdot]$ denotes channel-wise concatenation. This formulation lets the model explicitly distinguish reference content from generative constraints via the auxiliary mask channel. A modality-specific Context Projector processes each fused sequence $\mathbf{z}_{in, m}$. To preserve generative priors, projector weights for latent features are inherited from the original backbone embedders, while mask channel weights are initialized from scratch to accommodate the new modality inputs.

\noindent \textbf{Bypass Transformer Blocks.} For modality-specific control, we introduce independent visual and acoustic bypass branches comprising Bypass Transformer Blocks, strategically interleaved with even-numbered backbone layers. To inherit base model capabilities, internal Self-Attention, Text Cross-Attention, and FFN layers are initialized using pre-trained weights. We deliberately omit explicit audio-to-video and video-to-audio cross-attention mechanisms within these branches to prioritize specialized intra-modality refinement and maintain high computational efficiency. At each even-numbered layer $l$, the bypass branch processes context tokens, with the resulting output mapped by a dedicated projection layer to form a contextual hint for residual injection into the main backbone. This output projector is zero-initialized to ensure training stability by preventing initial disturbances to the frozen pre-trained priors.

\subsection{Inference Strategies for MMControl}

To address multi-modal synchronization, we introduce modality-specific scaling alongside a two-stage refinement process, ensuring high-fidelity generation and flexible control over individual streams.

\noindent \textbf{Modality-Specific Guidance Scaling.}
We propose Modality-Specific Guidance Scaling for decoupled control over distinct modalities. While bypass branches extract essential features, scaling factors $\gamma_v$ and $\gamma_a$ adaptively modulate their respective influence on joint generation for visual and acoustic streams. Specifically, in the $l$-th main transformer block, the hidden state $\mathbf{x}_{\text{main}}^{(l)}$ is updated via a gated residual connection that incorporates these contextual hints:
\begin{equation}
    \mathbf{x}_{\text{main}}^{(l)} = \text{MainBlock}^{(l)}(\mathbf{x}_{\text{main}}^{(l-1)}) + \gamma_m \cdot \text{Hint}_{m}^{(l)}, \quad m \in \{v, a\}.
\end{equation}
Here, $\text{Hint}_{m}^{(l)}$ denotes the modality-aware feature maps extracted from the corresponding $l$-th bypass block. Adjusting $\gamma_v$ and $\gamma_a$ during inference enables users to navigate the trade-off between strict conditional fidelity and generative creativity from pre-trained priors. For instance, a high $\gamma_v$ value ensures rigorous adherence to complex pose sequences, whereas a moderate $\gamma_a$ facilitates more natural prosody variations in synthesized speech.

\begin{figure*}[ht]
  \centering
  \includegraphics[width=0.95\textwidth]{figures/inference.pdf}
   \caption{
   \textbf{Two-Stage Progressive Inference Strategy.} Stage 1 forms a low-resolution ($h/2 \times w/2$) semantic base with modality-specific scales ($\gamma_v, \gamma_a$) for independent visual and acoustic guidance. Stage 2 applies distilled LoRA for fine-grained refinement and upscaling to full resolution ($h \times w$).
}
    \label{fig:inference_pipeline}
\end{figure*}

\noindent \textbf{Two-Stage Progressive Inference.} Following the established LTX-2~\cite{hacohen2026ltx} inference paradigm, we implement a Two-Stage Progressive Inference strategy to produce high-resolution outputs with optimal computational efficiency. This hierarchical framework explicitly decouples the initial formation of global semantic structures from subsequent refinement of fine-grained textures.

\textbf{Stage 1: Semantic Base Generation.} The model first generates a synchronized video and audio draft at a reduced spatial resolution ($h/2 \times w/2$). We use full MMCU conditions with a standard flow-matching schedule with Classifier-Free Guidance (CFG) to establish the foundational visual structure and acoustic timbre. During this phase, modality-specific scales $\gamma_v$ and $\gamma_a$ are employed to define the initial structural strength and cross-modal alignment.

\textbf{Stage 2: Progressive Detail Refinement.} This stage focuses on upscaling and refining the Stage 1 output to the target full resolution ($h \times w$). A dedicated spatial latent upsampler first processes the low-resolution video latents. To enhance fine-grained detail synthesis without full denoising overhead, we integrate a frozen, pre-trained Distilled LoRA from LTX-2 into the backbone transformer. 

During refinement, we apply a truncated noise schedule, typically spanning 3 steps, starting from the upsampled latents. CFG is disabled to preserve the established semantic layout, while MMCU signals are re-encoded at full resolution to provide precise guidance for high-frequency details. Finally, respective VAE decoders map these refined latents back into pixel space and high-fidelity audio waveforms.

\section{Experiments}

\subsection{Experimental Setup}

\noindent \textbf{Data Preparation.} \method is trained on a curated subset of 30,000 high-quality samples from Hallo3~\cite{cui2025hallo3}. The training objective covers three primary controllable tasks: generation conditioned on reference images and audio, reference images and depth maps, and reference images and poses, ensuring comprehensive multi-modal coverage. To maintain data integrity, we keep clips with a minimum duration of 2.0s, enabling non-overlapping reference-audio and target-video segments to prevent information leakage. During preprocessing, Grounded-SAM-2~\cite{ravi2024sam} precisely segments characters for visual reference images. For structural guidance, Depth-Anything-v2~\cite{yang2024depth} computes depth maps, and DWPose~\cite{yang2023effective} extracts whole-body poses. Qwen2.5-Omni~\cite{xu2025qwen2} generates detailed captions for structured textual prompts to bridge the modality gap.

\noindent \textbf{Implementation Details.}
\method is initialized from the pre-trained LTX-2 19B~\cite{hacohen2026ltx} backbone to leverage its generative knowledge. Optimization spans 7,200 steps using AdamW, with a peak learning rate of $1 \times 10^{-5}$ regulated by a cosine annealing scheduler for smooth convergence. We train on four NVIDIA H200 GPUs with a per-GPU batch size of 2 and two gradient accumulation steps. To enable flexible composition and robust classifier-free guidance, we drop both visual and acoustic control signals with probability 0.1. Training takes approximately 12 hours, highlighting the efficiency of our bypass-based optimization.

\noindent \textbf{Benchmark and Evaluation.} To assess \method, we construct a comprehensive evaluation benchmark with 200 samples for comparison with audio-driven joint generation and another 200 for depth-conditioned generation. For general video quality assessment, covering motion smoothness, aesthetic appeal, and temporal consistency, we follow the multidimensional VBench~\cite{huang2024vbench} protocol. Task-specific evaluations use Sync-C and Sync-D~\cite{chung2016out, prajwal2020lip} to measure audio-visual synchronization precision. To quantify semantic alignment, Text CLIP similarity~\cite{radford2021learning} is computed by averaging cosine similarity between the \texttt{[VISUAL]} portion of text prompts and eight uniformly sampled frames using a CLIP ViT-L/14 backbone. For identity preservation, Subject DINO similarity~\cite{caron2021emerging,liu2024grounding,ravi2024sam} is calculated between pre-segmented reference images and generated frames, leveraging a DINO ViT-B/16 model and Grounding-DINO with SAM2 for character isolation. We evaluate motion intensity via Dynamic Degree, which measures mean optical flow magnitude across 49 frames using RAFT-Large~\cite{teed2020raft}. For structural control, we calculate Mean Absolute Error (MAE) between input depth maps and those extracted from synthesized videos to determine spatial fidelity.
\subsection{Main Results}
\begin{table}[!htbp]
    \centering
    \footnotesize
    \caption{
        \textbf{Quantitative comparisons with prior methods.} We report sync (Sync-C, Sync-D) and visual metrics; \method leads on most metrics, showing superior generation quality and audio-visual sync.
    }
    \label{tab:main_metric}
    \resizebox{\textwidth}{!}{
    \begin{tabular}{lccccc|cc}
        \toprule
        \textbf{Method} & 
        \textbf{\shortstack{Text CLIP\\Similarity$\uparrow$}} & 
        \textbf{\shortstack{Subject DINO\\Similarity$\uparrow$}} & 
        \textbf{\shortstack{Dynamic\\Degree$\uparrow$}} & 
        \textbf{\shortstack{Aesthetic\\Quality$\uparrow$}} & 
        \textbf{\shortstack{Motion\\Smoothness$\uparrow$}} &
        \textbf{\shortstack{Sync-D$\downarrow$}} &
        \textbf{\shortstack{Sync-C$\uparrow$}} \\
        \midrule
        SadTalker~\cite{zhang2023sadtalker}    & 0.2410 & 0.8752 & 0.0057 & 0.5434 & \textbf{0.9977} & 10.696 & 2.474 \\
        AniPortrait~\cite{wei2024aniportrait}  & 0.2302 & 0.8857 & 0.0368 & 0.5350 & 0.9965          & 12.383 & 1.322 \\
        HunyuanCustom                          & 0.2203 & 0.7816 & 0.6150 & 0.5003 & 0.9955          & 10.912 & 1.628 \\
        Hallo3~\cite{cui2025hallo3}              & 0.2337 & 0.8873 & \textbf{0.9184} & 0.5169 & 0.9667 & \textbf{10.243} & 2.550 \\
        
        \midrule
        \textbf{\method~(Ours)}                & \textbf{0.2546} & \textbf{0.8948} & 0.5750 & \textbf{0.5461} & 0.9954 & 10.506 & \textbf{2.716} \\
        \bottomrule
    \end{tabular}
    }
\end{table}

\noindent \textbf{Comparison with Audio-Driven Baselines.} Since customization frameworks such as Hallo3~\cite{cui2025hallo3}, SadTalker~\cite{zhang2023sadtalker}, and AniPortrait~\cite{wei2024aniportrait} lack integrated audio-video synthesis, we benchmark them against \method. Following MoCha~\cite{wei2025mocha}, we use the ground-truth first frame as a shared reference for all models, maintaining visual consistency. While baselines rely on ground-truth (GT) audio and first frame, \method must jointly generate speech and video from text, a more complex generative task. Table~\ref{tab:main_metric} shows \method achieves the highest Sync-C score despite the baselines' GT audio advantage, proving our joint DiT captures inherent audio-visual correlations. \method leads in Text CLIP and Subject DINO similarity, confirming superior semantic alignment and identity preservation. Although Hallo3 exhibits a higher Dynamic Degree, \method better balances motion intensity and visual aesthetics, yielding superior appeal.

\noindent \textbf{Structural Depth Control.}
We use depth maps as representative signals for structural control. As reported in Table~\ref{tab:depth_metric}, \method achieves a Mean MAE ($\times 100$) of \textbf{4.52}, substantially improving over VideoComposer (15.41) and recent SOTA VACE (5.35). Beyond structural precision, \method yields superior performance across semantic and motion metrics, especially subject similarity and dynamic degree. These results show that the Multi-Modal Control Unit (MMCU) preserves fine-grained spatial structures while maintaining high visual-semantic coherence during joint diffusion.

\begin{table}[!htbp]
    \centering
    \footnotesize
    \caption{
        \textbf{Quantitative comparisons on structural depth control.} We report visual metrics and depth MAE between input and generated videos. \method leads on most metrics and has the lowest structural error.
    }
    \label{tab:depth_metric}
    \resizebox{\textwidth}{!}{
    \begin{tabular}{lccccc|c}
        \toprule
        \textbf{Method} & 
        \textbf{\shortstack{Text CLIP\\Similarity$\uparrow$}} & 
        \textbf{\shortstack{Subject DINO\\Similarity$\uparrow$}} & 
        \textbf{\shortstack{Aesthetic\\Quality$\uparrow$}} & 
        \textbf{\shortstack{Dynamic\\Degree$\uparrow$}} & 
        \textbf{\shortstack{Motion\\Smoothness$\uparrow$}} &
        \textbf{\shortstack{Mean MAE\\$\times 100 \downarrow$}} \\
        \midrule
        VideoComposer~\cite{wang2023videocomposer} & 0.2252          & 0.5888          & 0.4535          & 0.5900          & 0.9667          & 15.41 \\
        ControlVideo~\cite{zhang2024controlvideo}   & 0.2450          & 0.6698          & 0.5652          & 0.2450          & 0.9852          & 10.64 \\
        VACE~\cite{jiang2025vace}                   & 0.2434          & 0.8894          & \textbf{0.5722} & 0.4100          & \textbf{0.9952} &  5.35 \\
        
        \midrule
        \textbf{\method~(Ours)}                     & \textbf{0.2453} & \textbf{0.8982} & 0.5556          & \textbf{0.6650} & \textbf{0.9952} & \textbf{4.52} \\
        \bottomrule
    \end{tabular}
    }
\end{table}

\noindent \textbf{Structural Pose Control.}
As reported in Table~\ref{tab:pose_metric}, \method achieves the lowest pose MAE ($\times 100$) of \textbf{3.07}, outperforming VACE (3.29), Text2Video-Zero (6.40), and ControlVideo (6.59). It further obtains the best temporal consistency and motion smoothness, indicating the proposed control mechanism improves pose adherence while preserving stable temporal dynamics.

\begin{table}[!htbp]
    \centering
    \footnotesize
    \caption{
        \textbf{Quantitative comparisons on structural pose control.} We report visual metrics and pose MAE between input and generated videos. \method achieves the lowest pose error and best temporal consistency.
    }
    \label{tab:pose_metric}
    \resizebox{\textwidth}{!}{
    \begin{tabular}{lcccc|c}
        \toprule
        \textbf{Method} &
        \textbf{\shortstack{Subject DINO\\Similarity$\uparrow$}} &
        \textbf{\shortstack{Dynamic\\Degree$\uparrow$}} &
        \textbf{\shortstack{Temporal\\Consistency$\uparrow$}} &
        \textbf{\shortstack{Motion\\Smoothness$\uparrow$}} &
        \textbf{\shortstack{Mean MAE\\$\times 100 \downarrow$}} \\
        \midrule
        Text2Video-Zero~\cite{khachatryan2023text2video} & 0.9823          & \textbf{0.6750} & 0.9682 & 0.9734 & 6.40 \\
        ControlVideo~\cite{zhang2024controlvideo}        & \textbf{0.9842} & 0.2400          & 0.9823 & 0.9865 & 6.59 \\
        VACE~\cite{jiang2025vace}                        & 0.9639          & 0.3800          & 0.9897 & 0.9946 & 3.29 \\
        \midrule
        \textbf{\method~(Ours)}                          & 0.9729          & 0.5250          & \textbf{0.9906} & \textbf{0.9952} & \textbf{3.07} \\
        \bottomrule
    \end{tabular}
    }
\end{table}

\noindent \textbf{Human and Audio Evaluation.}
We provide human evaluation and generated-audio quality analysis in the supplementary material. The user study shows \method obtains the highest overall score (3.58) across six perceptual criteria, while the audio evaluation reports speaker similarity (SIM-o 0.2574), naturalness (UTMOS 3.49), and transcription accuracy (7.96\% weighted WER).

\subsection{Qualitative Comparisons}

\begin{figure}[ht]
    \centering
    \includegraphics[width=1\textwidth]{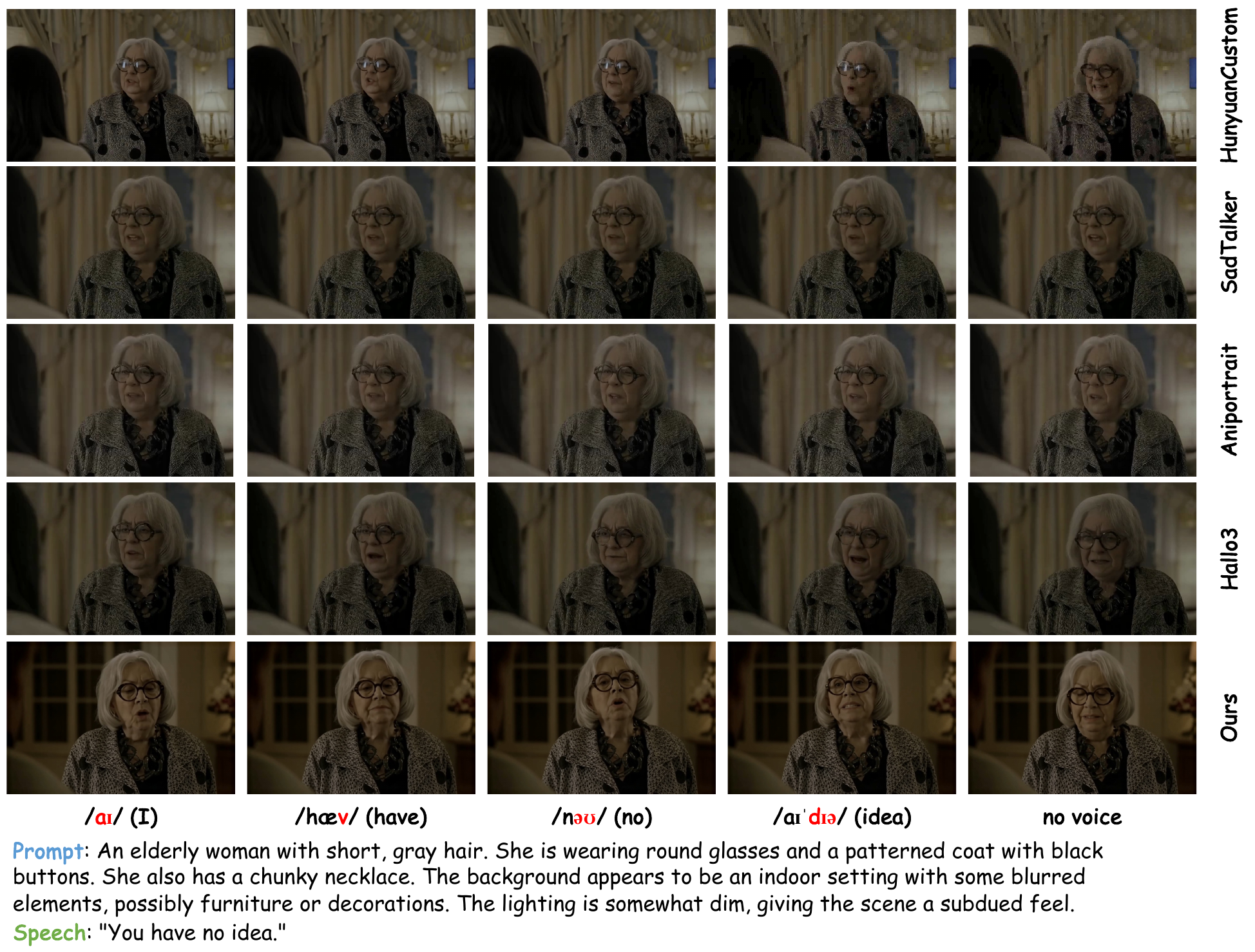}
    \caption{
        \textbf{Qualitative comparison with baseline methods.} 
        As shown, baseline approaches such as HunyuanCustom, SadTalker, Hallo3 and AniPortrait often produce noticeable facial artifacts and largely static backgrounds, which limits scene expressiveness. 
        \method generates high-fidelity videos with richer motion and natural transitions while maintaining stable character ID and precise audio-visual synchronization.
    }
    \label{fig:result_comparison}
    \vspace{-6mm}
\end{figure}

\noindent \textbf{Comparison with Audio-Driven Baselines.} As evidenced in \cref{fig:result_comparison}, baseline methods such as AniPortrait and Hallo3 frequently exhibit noticeable facial artifacts and temporal jitters, especially during complex motion transitions. Furthermore, the backgrounds generated by these modular approaches remain largely static, an inherent limitation that suppresses scene dynamics and restricts overall narrative expressiveness. In contrast, \method produces high-fidelity videos with more nuanced motion and naturalistic temporal coherence that more faithfully align with the narrative prompts. Additional qualitative results are provided in the supplementary material to further demonstrate the robustness of our framework.

\begin{figure}[ht]
    \centering
    \includegraphics[width=1\textwidth]{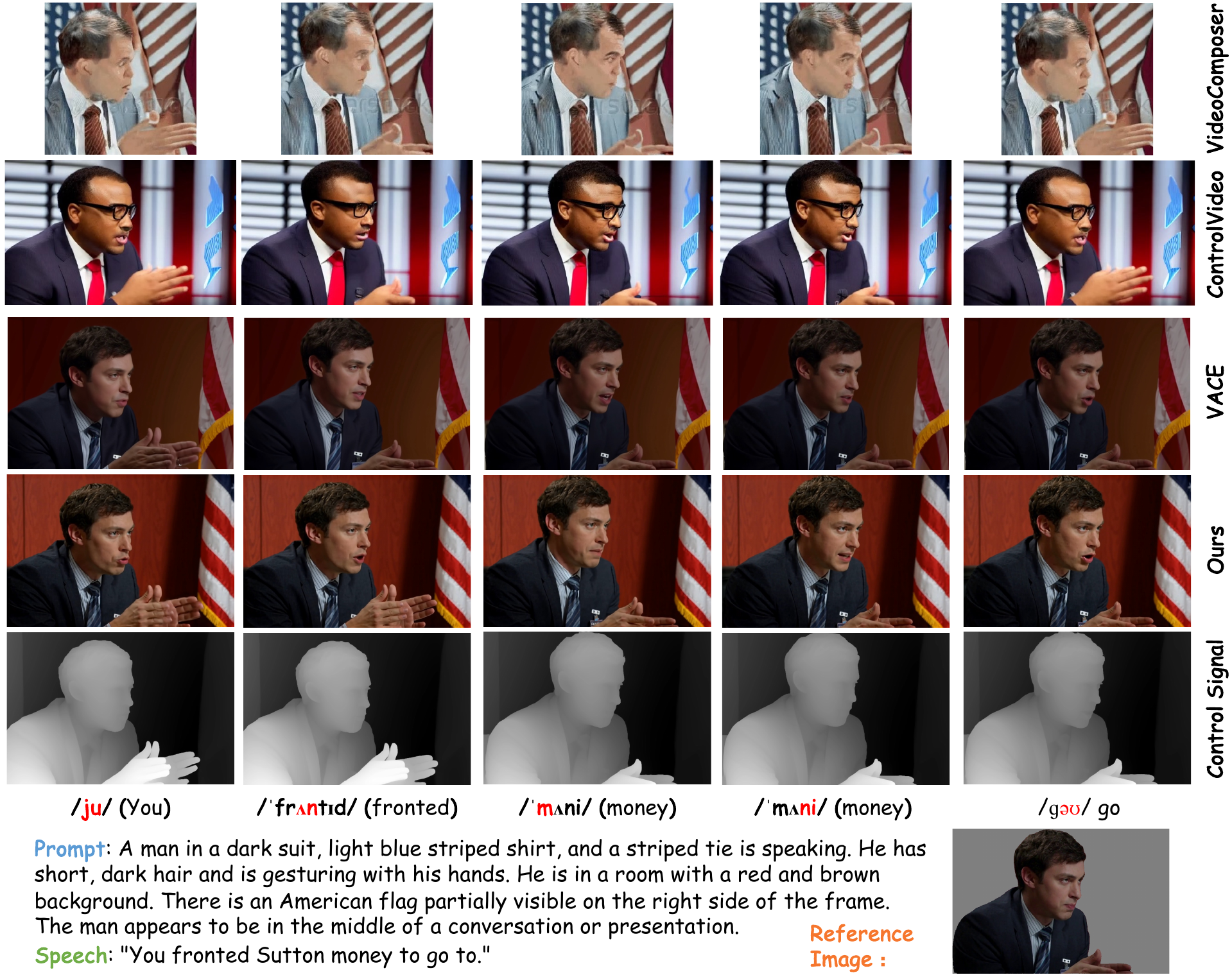}
    \caption{
        \textbf{Qualitative comparison of depth control.} The visual quality of images from VideoComposer and ControlVideo is lower. VACE shows weaker adherence to fine-grained depth cues (\eg, hand gestures). Our method provides high visual quality and precise control adherence.
    }
    \label{fig:result_depth}
\end{figure}

\noindent \textbf{Comparison with Depth Control Baselines.} The effectiveness of depth-control generation is further evaluated in \cref{fig:result_depth}. Compared to VideoComposer and ControlVideo, our approach exhibits significantly higher visual fidelity and subject consistency, avoiding the common identity-shifting artifacts and blurred textures seen in these baselines. While VACE produces high-quality frames, it shows weaker adherence to the intricate spatial structures and hand gestures provided in the depth control signal. In contrast, \method demonstrates stronger structural alignment, faithfully reconstructing complex motions while preserving the reference identity. This comparison focuses on visual structural adherence under the same depth input. Beyond this visual-control setting, \method further supports joint audio-video synthesis, enabling synchronized speech and lip motion within the same generation process.

\begin{figure}[ht]
    \centering
    \includegraphics[width=1\textwidth]{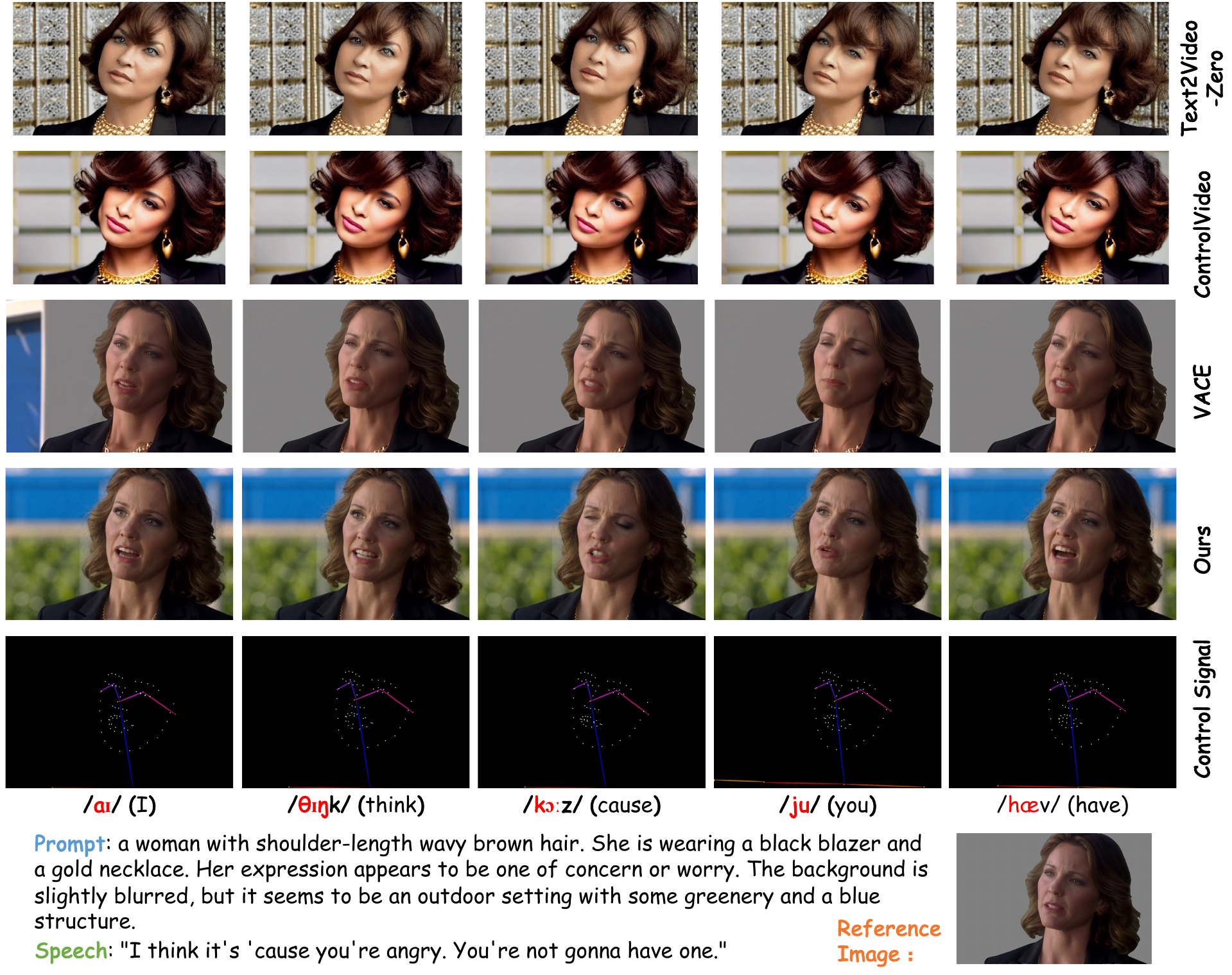}
    \caption{
        \textbf{Our method better adheres to structural pose signals and text prompts.} Compared to baselines, \method generates videos with higher visual quality, more accurate identity, and precise audio-visual synchronization.
    }
    \label{fig:result_pose}
\end{figure}

\noindent \textbf{Comparison with Pose Control Baselines.} The efficacy of pose-conditioned generation is qualitatively evaluated in \cref{fig:result_pose}. Compared to Text2Video-Zero and ControlVideo, the proposed approach shows superior visual fidelity and stronger adherence to the textual prompt, including the character's attire and outdoor background. While VACE achieves relatively high image quality, it is less consistent with the pose-conditioned scene composition in \cref{fig:result_pose}, including the specified background and character layout. In contrast, \method preserves the reference identity, follows the pose guidance, and generates synchronized audio in the same process, yielding a more coherent multi-modal output.

\begin{figure}[ht]
    \centering
    \includegraphics[width=1\textwidth]{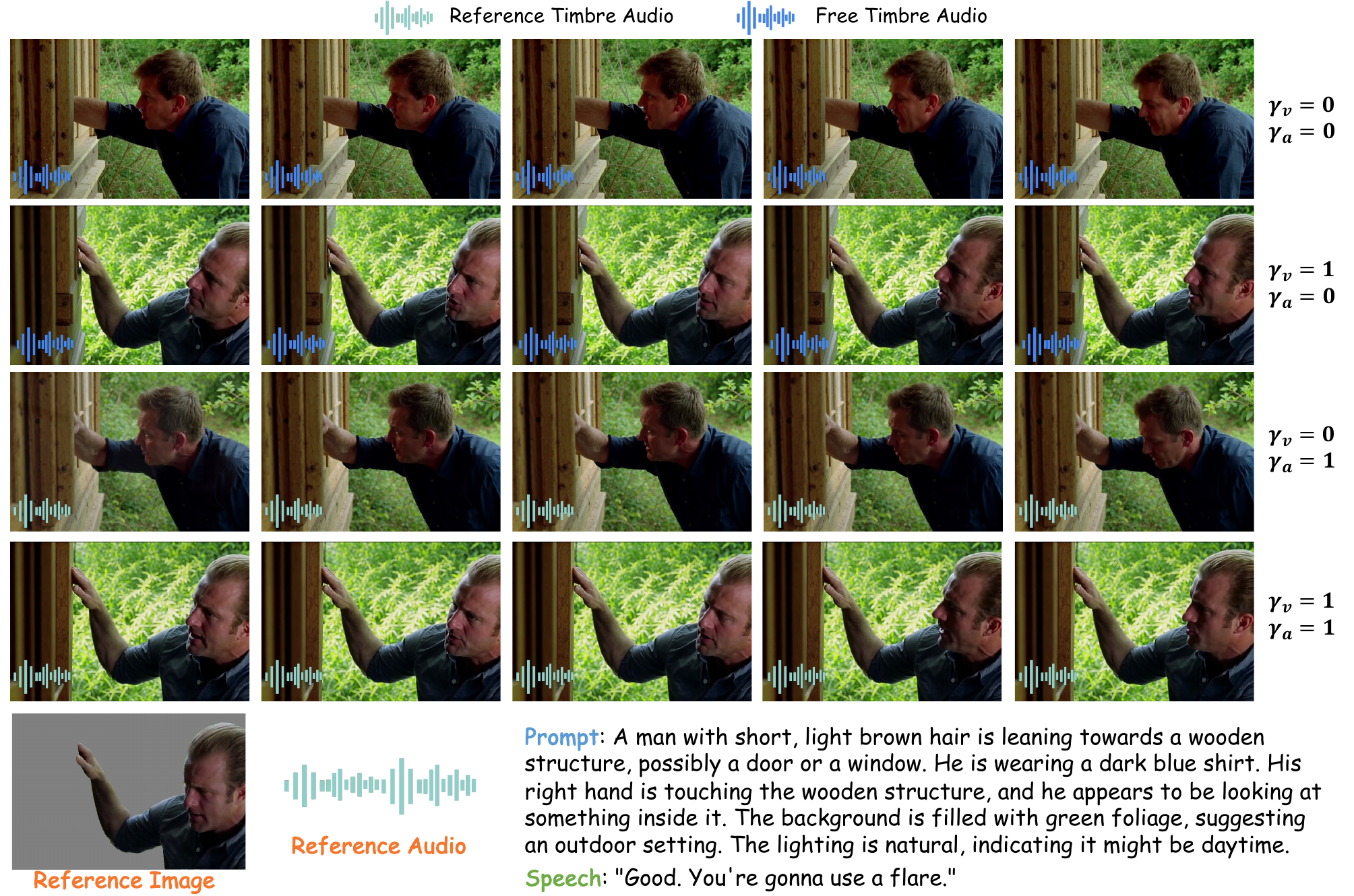}
    \caption{
        \textbf{Modality-Specific Guidance Scaling.} Varying $\gamma_v$ and $\gamma_a$ independently controls visual identity and acoustic timbre. With $\gamma_v=1$, identity follows the reference image; with $\gamma_a=1$, speech follows the reference voice.
    }
    \label{fig:scaling_study}
\end{figure}

\noindent \textbf{Decoupled Control Study.} To assess independent control, we qualitatively vary the modality-specific scaling factors $\gamma_v$ and $\gamma_a$ during inference. In \cref{fig:scaling_study}, when $\gamma_v=0$ and $\gamma_a=0$, the model generates a generic character and default voice guided exclusively by the textual prompt, reflecting the raw generative prior. With $\gamma_v=1$, \method anchors the character's visual identity to the reference image while the acoustic timbre remains unconditioned. Conversely, with $\gamma_a=1$ and $\gamma_v=0$, the model preserves the reference voice timbre but synthesizes a novel visual persona. The full configuration ($\gamma_v=1, \gamma_a=1$) achieves precise synchronization of both identity and timbre. This decoupled control capability lets users flexibly navigate the trade-off between reference fidelity and generative creativity across distinct modalities.

\section{Conclusion and Future Work}


We study Multi-Modal Controllable Joint Generation for synchronized audio-video synthesis guided by textual, reference, and structural signals. MMControl is a unified framework built on a joint Diffusion Transformer. It uses the Multi-Modal Control Unit (MMCU) to harmonize inputs and a Dual-Stream Bypass to inject modality-specific hints into a frozen generative backbone. Modality-Specific Guidance Scaling enables independent control over visual identity and acoustic timbre during inference. Quantitative and qualitative results show that MMControl achieves state-of-the-art audio-visual synchronization and structural adherence to depth and pose constraints.

Despite promising results, challenges remain. Future work includes extending the framework to multi-character or conversational settings requiring complex cross-speaker synchronization and long-term temporal coherence. We also aim to explore end-to-end systems adapting to diverse control modalities and unseen speech styles without task-specific fine-tuning. This work lays a foundation for unified, user-centric multi-modal generation for personalized media creation.



\newpage
{
\bibliographystyle{splncs04}
\bibliography{main}
}

\newpage
\appendix
\renewcommand\thesection{\Alph{section}}
\renewcommand\thefigure{S\arabic{figure}}
\renewcommand\thetable{S\arabic{table}}
\renewcommand\theequation{S\arabic{equation}}

\renewcommand\theHsection{appendix.\Alph{section}}
\renewcommand\theHfigure{S.\arabic{figure}}
\renewcommand\theHtable{S.\arabic{table}}
\renewcommand\theHequation{S.\arabic{equation}}

\setcounter{figure}{0}
\setcounter{table}{0}
\setcounter{equation}{0}

\section*{Appendix Overview}

This appendix provides additional implementation details, empirical analysis, and extended results to supplement the main paper. It is organized as follows:
\begin{itemize}
    \item \textbf{\cref{sec:app_human_eval}: Human Evaluation} \\
    Provides further details on subjective human evaluation.

    \item \textbf{\cref{sec:app_audio_quality}: Audio Quality Evaluation} \\
    Provides evaluation metrics and quantitative results for generated audio quality.

    \item \textbf{\cref{sec:app_ablation}: Ablation} \\
    Provides component and guidance-scaling ablations.

    \item \textbf{\cref{sec:app_avcontrol}: Comparison with AVControl} \\
    Provides an additional comparison with the concurrent AVControl framework.

\end{itemize}

\section{Human Evaluation}
\label{sec:app_human_eval}

We perform a user study to subjectively evaluate our method on reference-image-conditioned joint audio-video generation. In this setting, MMControl takes only a reference image as input to jointly generate audio and video, while baseline methods use ground-truth audio and the first video frame. Participants rated each method on six criteria using a 4-point scale: 1 = Not Good, 2 = Borderline Reject, 3 = Borderline Accept, 4 = Good.

\begin{itemize}
    \item \textbf{Lip-Sync Accuracy}: How well the character's lip movements match the generated audio.
    \item \textbf{Facial Expression Realism}: Whether the character's facial expressions look natural and are contextually appropriate.
    \item \textbf{Action Naturalness}: How natural the character's body movements and gestures appear in relation to the audio.
    \item \textbf{Text Alignment}: How well the character's behaviors match the descriptions in the text prompt.
    \item \textbf{Subject Alignment}: Whether the generated character is consistent with the identity and appearance of the reference subject.
    \item \textbf{Visual Quality}: Overall visual fidelity of the output, including the presence of artifacts or other defects.
\end{itemize}

\begin{table*}[t]
    \centering
    \footnotesize
    \caption{
    \textbf{Human evaluation results.}
    Each criterion was rated on a 4-point Likert scale by 20 independent participants. Participants rated each method on six aspects: lip-sync accuracy, facial expression realism, action naturalness, text alignment, subject alignment, and visual quality.
    }
    \label{tab:human_eval}
    \setlength{\tabcolsep}{6pt}
    \resizebox{\textwidth}{!}{
    \begin{tabular}{lcccccc c}
        \toprule
        \textbf{Method} &
        \textbf{\shortstack{Lip-Sync\\Accuracy}} &
        \textbf{\shortstack{Facial Expression\\Realism}} &
        \textbf{\shortstack{Action\\Naturalness}} &
        \textbf{\shortstack{Text\\Alignment}} &
        \textbf{\shortstack{Subject\\Alignment}} &
        \textbf{\shortstack{Visual\\Quality}} &
        \textbf{Overall} \\
        \midrule
        Hallo3~\cite{cui2025hallo3}        & 3.07 & 3.11 & 3.08 & 3.28 & 3.53 & 3.22 & 3.22 \\
        HunyuanCustom~\cite{hu2025hunyuancustom} & 2.90 & 3.17 & 3.21 & 3.18 & 3.38 & 3.39 & 3.20 \\
        SadTalker~\cite{zhang2023sadtalker} & 2.35 & 1.78 & 1.78 & 2.78 & 3.31 & 2.68 & 2.45 \\
        AniPortrait~\cite{wei2024aniportrait} & 1.53 & 1.64 & 1.58 & 2.56 & 3.15 & 2.37 & 2.14 \\
        \midrule
        MMControl (Ours) & \textbf{3.41} & \textbf{3.60} & \textbf{3.58} & \textbf{3.57} & \textbf{3.56} & \textbf{3.77} & \textbf{3.58} \\
        \bottomrule
    \end{tabular}
    }
\end{table*}

To ensure robust evaluation, we randomly sampled instances from the test set and recruited 20 independent participants to evaluate each sample, with each participant rating 5 data points per method. As summarized in \cref{tab:human_eval}, \method consistently outperforms all baseline approaches across all six criteria, achieving an overall average score of \textbf{3.58}, which surpasses the best-performing baseline by a notable margin.

Specifically, \method achieves the highest \textbf{Lip-Sync Accuracy} (3.41), demonstrating tighter lip--speech alignment even against methods using ground-truth audio. For \textbf{Facial Expression Realism} (3.60) and \textbf{Action Naturalness} (3.58), \method outperforms all baselines by a substantial margin, reflecting more natural character behaviors. On \textbf{Text Alignment} (3.57), our model better follows the semantic intent of the prompt, benefiting from the structured text condition in MMCU. For \textbf{Subject Alignment} (3.56), reference-image conditioning ensures superior identity consistency throughout generation. Finally, the highest \textbf{Visual Quality} score (3.77) confirms our outputs are largely free of artifacts, outperforming other methods by a large margin. These results collectively validate the effectiveness of \method in producing compelling, identity-faithful, and well-synchronized audio-visual content.

\section{Audio Quality Evaluation}
\label{sec:app_audio_quality}

We evaluate the generated audio quality using three standard metrics that capture speaker similarity, speech naturalness, and transcription accuracy. To ensure a fair and meaningful evaluation, we select 150 test samples that have both a reference image condition and a reference audio condition, which corresponds to the full multi-modal control setting of \method. \cref{tab:app_audio_quality} reports the quantitative results. The evaluation protocol follows the official ZipVoice~\cite{zhu2025zipvoice} implementation.

\begin{itemize}
    \item \textbf{Speaker Similarity (SIM-o)}: Evaluates whether the generated voice preserves the identity of the reference speaker. Speaker embeddings are extracted using an ECAPA-TDNN~\cite{desplanques2020ecapa} verifier with a WavLM-Large~\cite{chen2022wavlm} front-end, and similarity is measured by cosine distance. A higher score indicates better speaker identity preservation.
    \item \textbf{UTMOS (Objective MOS)}: An automated speech quality estimator based on UTMOS22Strong~\cite{saeki2022utmos}, which predicts perceptual naturalness scores from 16\,kHz audio without human listeners. Higher scores indicate better perceived quality.
    \item \textbf{Word Error Rate (WER)}: Computed with the Whisper-large-v3~\cite{radford2023robust} ASR model following the Seed-TTS~\cite{anastassiou2024seed} protocol. We report both the weighted WER and the SeedTTS average WER. Lower values are better.
\end{itemize}

\begin{table}[t]
    \centering
    \caption{Audio quality evaluation results on 150 sampled test cases. Higher is better ($\uparrow$) for SIM-o and UTMOS; lower is better ($\downarrow$) for WER.}
    \label{tab:app_audio_quality}
    \setlength{\tabcolsep}{12pt}
    \begin{tabular}{l c}
        \toprule
        \textbf{Metric} & \textbf{Score} \\
        \midrule
        Speaker Similarity (SIM-o) $\uparrow$ & 0.2574 \\
        UTMOS $\uparrow$                       & 3.49 \\
        WER -- weighted $\downarrow$           & 7.96\% \\
        WER -- SeedTTS avg $\downarrow$        & 17.58\% \\
        \bottomrule
    \end{tabular}
\end{table}

\section{Ablation}
\label{sec:app_ablation}

We conduct ablation studies to evaluate the contributions of MMCU, the Dual-Stream Bypass design, and Modality-Specific Guidance Scaling. The component variants are: M1 removes masks from MMCU; M2 removes projector weight inheritance; B1 replaces the dual-stream design with unified attention; B2 removes the audio bypass; B3 removes the visual bypass; and B4 adds audio-video cross-attention inside bypass branches.

\begin{table}[!hbp]
    \centering
    \scriptsize
    \caption{
        \textbf{Component ablation.} We evaluate each variant on 90 paired samples.
    }
    \label{tab:app_ablation_components}
    \setlength{\tabcolsep}{2.2pt}
    \renewcommand{\arraystretch}{0.88}
    \begin{adjustbox}{max width=0.92\textwidth}
    \begin{tabular}{llccccccc}
        \toprule
        & & & \multicolumn{2}{c}{\textbf{MMCU}} & \multicolumn{4}{c}{\textbf{Dual-Stream}} \\
        \cmidrule(lr){4-5}\cmidrule(lr){6-9}
        Task & Metric & Full & M1 & M2 & B1 & B2 & B3 & B4 \\
        \midrule
        \multirow{3}{*}{Audio}
        & Sync-C $\uparrow$        & \textbf{2.73} & 2.53 & 2.36 & 2.15 & 2.20 & 2.15 & 2.73 \\
        & SIM-o $\uparrow$         & \textbf{0.22} & 0.17 & 0.16 & 0.17 & 0.10 & 0.11 & 0.14 \\
        & Dyn.\,Deg $\uparrow$     & \textbf{0.67} & 0.33 & 0.13 & \textbf{0.67} & 0.43 & 0.33 & 0.60 \\
        \midrule
        \multirow{2}{*}{Depth}
        & MAE $\times 10^{-2}$ $\downarrow$ & \textbf{4.65} & 5.01 & 18.32 & 5.13 & 4.97 & 19.35 & 4.87 \\
        & Dyn.\,Deg $\uparrow$     & 0.67 & \textbf{0.70} & 0.13 & \textbf{0.70} & 0.60 & 0.20 & 0.67 \\
        \midrule
        \multirow{2}{*}{Pose}
        & MAE $\times 10^{-2}$ $\downarrow$ & \textbf{3.78} & 3.92 & 14.42 & 4.00 & 4.24 & 16.83 & 4.01 \\
        & Dyn.\,Deg $\uparrow$     & 0.43 & \textbf{0.57} & 0.27 & 0.50 & 0.43 & 0.30 & 0.43 \\
        \bottomrule
    \end{tabular}
    \end{adjustbox}
\end{table}

As shown in \cref{tab:app_ablation_components}, removing MMCU masks degrades audio synchronization, with Sync-C dropping from 2.73 to 2.53, while removing projector weight inheritance severely hurts structural control, increasing depth MAE from 4.65 to 18.32 and pose MAE from 3.78 to 14.42. The dual-stream design is also important: unified attention reduces Sync-C to 2.15, removing the audio bypass lowers SIM-o from 0.22 to 0.10, and removing the visual bypass substantially increases depth and pose MAE. Adding audio-video cross-attention inside the bypass branches also reduces speaker similarity, which supports our skip-cross-attention design.

\begin{table}[H]
    \centering
    \scriptsize
    \caption{
        \textbf{Guidance scaling ablation.} We vary $\gamma_v$ and $\gamma_a$ independently.
    }
    \label{tab:app_ablation_guidance}
    \setlength{\tabcolsep}{5pt}
    \renewcommand{\arraystretch}{0.9}
    \begin{tabular}{lcccccc}
        \toprule
        & \multicolumn{3}{c}{$\gamma_v$ ($\gamma_a{=}1$)} & \multicolumn{3}{c}{$\gamma_a$ ($\gamma_v{=}1$)} \\
        \cmidrule(lr){2-4}\cmidrule(lr){5-7}
        \textbf{Metric} & \textbf{0.5} & \textbf{1.0} & \textbf{1.3} & \textbf{0.5} & \textbf{1.0} & \textbf{1.3} \\
        \midrule
        Subj.\,DINO $\uparrow$ & 0.860 & 0.896 & \textbf{0.911} & 0.894 & 0.896 & 0.906 \\
        SIM-o $\uparrow$       & 0.16  & \textbf{0.22} & 0.16    & 0.16  & \textbf{0.22} & 0.14  \\
        \bottomrule
    \end{tabular}
\end{table}

\cref{tab:app_ablation_guidance} further studies modality-specific guidance scaling. Increasing $\gamma_v$ to 1.3 improves subject preservation, raising Subject DINO from 0.860 to 0.911, while changing $\gamma_a$ mainly affects audio adherence. For example, reducing $\gamma_a$ to 0.5 relaxes acoustic reference strength and lowers SIM-o from 0.22 to 0.16. These results show that the guidance scales provide inference-time control over visual identity and acoustic timbre without retraining.

\section{Comparison with AVControl}
\label{sec:app_avcontrol}

We additionally compare \method with the concurrent AVControl framework~\cite{ben2026avcontrol}, which trains task-specific IC-LoRA adapters for different control modalities. Since AVControl's audio control focuses on loudness while \method targets reference voice timbre, we compare the two methods on shared structural controls, namely depth and pose. As shown in \cref{tab:app_avcontrol}, \method achieves substantially lower MAE for both depth and pose control, while also improving subject preservation and motion smoothness. AVControl obtains a slightly higher aesthetic score in the pose setting, whereas \method provides stronger structural adherence and supports joint audio-video generation in the same framework.

\begin{table}[H]
    \centering
    \footnotesize
    \caption{
        \textbf{Comparison with AVControl.} We compare depth and pose control performance on 60 samples. Lower is better for MAE; higher is better for all other metrics.
    }
    \label{tab:app_avcontrol}
    \resizebox{\textwidth}{!}{
    \begin{tabular}{llcccc}
        \toprule
        \textbf{Task} &
        \textbf{Method} &
        \textbf{\shortstack{Mean MAE\\$\times 100 \downarrow$}} &
        \textbf{\shortstack{Subject DINO\\Similarity$\uparrow$}} &
        \textbf{\shortstack{Motion\\Smoothness$\uparrow$}} &
        \textbf{\shortstack{Aesthetic\\Quality$\uparrow$}} \\
        \midrule
        \multirow{2}{*}{Depth}
        & AVControl~\cite{ben2026avcontrol} & 11.49 & 0.851 & 0.991 & 0.527 \\
        & \textbf{\method~(Ours)}                & \textbf{4.52} & \textbf{0.898} & \textbf{0.995} & \textbf{0.556} \\
        \midrule
        \multirow{2}{*}{Pose}
        & AVControl~\cite{ben2026avcontrol} & 7.57 & 0.842 & 0.992 & \textbf{0.569} \\
        & \textbf{\method~(Ours)}                & \textbf{3.07} & \textbf{0.973} & \textbf{0.995} & 0.549 \\
        \bottomrule
    \end{tabular}
    }
\end{table}

\end{document}